\documentclass[Afour,sageh,times]{sagej}
\usepackage{moreverb,url}

\usepackage{subfigure} 
\usepackage[colorlinks,bookmarksopen,bookmarksnumbered,citecolor=red,urlcolor=red]{hyperref}

\newcommand\BibTeX{{\rmfamily B\kern-.05em \textsc{i\kern-.025em b}\kern-.08em
T\kern-.1667em\lower.7ex\hbox{E}\kern-.125emX}}

\def\volumeyear{2016}

\begin{document}

\runninghead{Sharma, Huang, Nair, Wen, Petlowany, Moore, Wanna and Pryor}

\title{The Collection of a Human Robot Collaboration Dataset for Cooperative Assembly in Glovebox Environments}

\author{Shivansh Sharma\affilnum{1}, Mathew Huang\affilnum{1}, Sanat Nair\affilnum{1}, Alan Wen\affilnum{1}, Christina Petlowany\affilnum{1}, Juston Moore\affilnum{2}, Selma Wanna\affilnum{1,2}, and Mitch Pryor\affilnum{1}}

\affiliation{\affilnum{1}Nuclear and Applied Robotics Group, Department of Mechanical Engineering\\
\affilnum{2}Advanced Research in Cyber Systems, Los Alamos National Laboratory.}

\corrauth{Selma Wanna, Los Alamos National Laboratory,
Los Alamos, USA.}

\email{slwanna@lanl.gov}

\begin{abstract}
Industry 4.0 introduced AI as a transformative solution for modernizing manufacturing processes. Its successor, Industry 5.0, envisions humans as collaborators and experts guiding these AI-driven manufacturing solutions. Developing these techniques necessitates algorithms capable of safe, real-time identification of human positions in a scene, particularly their hands, during collaborative assembly. Although substantial efforts have curated datasets for hand segmentation, most focus on residential or commercial domains. Existing datasets targeting industrial settings predominantly rely on synthetic data, which we demonstrate does not effectively transfer to real-world operations. Moreover, these datasets lack uncertainty estimations critical for safe collaboration. Addressing these gaps, we present HAGS: Hand and Glove Segmentation Dataset. This dataset provides challenging examples to build applications toward hand and glove segmentation in industrial human-robot collaboration scenarios as well as assess out-of-distribution images, constructed via green screen augmentations, to determine ML-classifier robustness. We study state-of-the-art, real-time segmentation models to evaluate existing methods. Our \href{https://dataverse.tdl.org/dataset.xhtml?persistentId=doi:10.18738/T8/85R7KQ}{dataset} and \href{https://github.com/UTNuclearRoboticsPublic/assembly\_glovebox\_dataset}{baselines} are publicly available.
\end{abstract}


\maketitle

\section{Introduction}
\label{sec:introduction}

\begin{figure*}[htbp]
    \centering
    \begin{minipage}{0.45\textwidth}
        \centering
        \subfigure[]{
            \includegraphics[width=\textwidth]{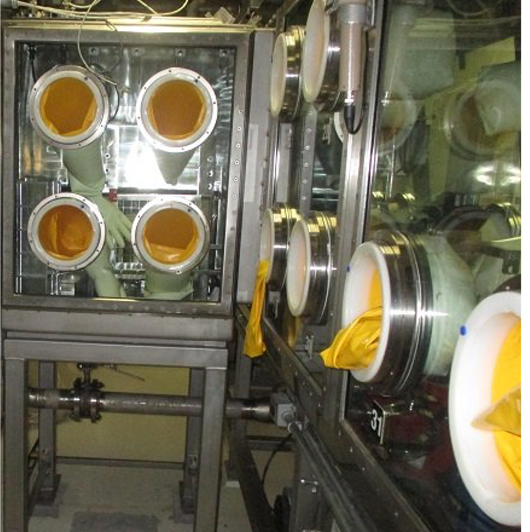}
            \label{fig:subfig_a}
        }
    \end{minipage}%
    \begin{minipage}{0.51\textwidth}
        \centering
        \subfigure[]{
            \includegraphics[width=\textwidth]{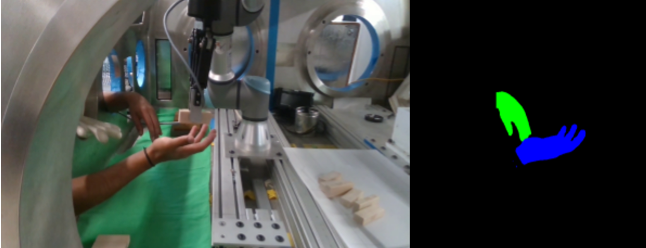}
            \label{fig:subfig_b}
        }\\
        \subfigure[]{
            \includegraphics[width=\textwidth]{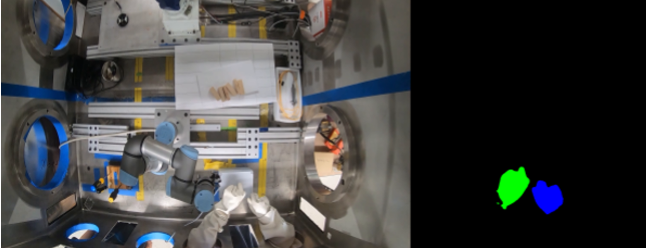}
            \label{fig:subfig_c}
        }
    \end{minipage}
    \caption{Left: (a) Example of a glovebox suite from \citep{gb_srs_pic}. Right: (b,c) Examples of RGB images from the HAGs dataset \citep{HAGS} with corresponding pixel-wise labels.}
    \label{fig:gb-intro}
\end{figure*}

Gloveboxes are self-contained spaces that allow workers to handle hazardous materials using gloves affixed to sealed portholes (see Figure \ref{fig:gb-intro}.) This setup protects operators from exposure and prevents unfiltered material releases into the environment \citep{gb_defn}. Workers and researchers that handle hazardous materials in gloveboxes face issues such as ergonomic injuries and potential hazardous exposure via glove tear. The use of robots in these environments can mitigate many of these problems. This paper seeks to improve glovebox manufacturing techniques for applications involving Special Nuclear Materials (SNMs) \citep{bean} to minimize radiation exposure to personnel. SNM handing is important; for example the production of Mo-99 decay products, used in the majority of medical diagnostic procedures, involves the use of highly enriched uranium \citep{mediso}. SNM manufacturing applications include medical isotope production \citep{mediso}, DoD applications \citep{joshwills}, radiological imaging \citep{nickhash}, disposition of spent nuclear fuel \citep{wms}, and handling irradiated samples for neutron activation analysis \citep{jhash}. Current manufacturing procedures potentially expose personnel to extremely high radiation doses, and robotics has the potential to reduce the time and increase the distance between radioactive sources and personnel. Ideally applications are fully automated, but niche applications and experimental activities related to handling SNMs will likely require humans to remain - at least partially - in the loop. The safety standards a robotic system must meet in all of these applications is exceptionally stringent.

Recent successes in the machine learning community have inspired robotics researchers to develop large-scale datasets aimed at achieving breakthroughs similar to ImageNet \citep{imagenet} for robotic research \citep{brohan2023rt1, open_x_embodiment_rt_x_2023}. While there are innumerable possibilities for robotics applications, previous datasets prioritize instruction following in residential environments \citep{brohan2023rt1, open_x_embodiment_rt_x_2023}. Unfortunately, these datasets often overlook elements of Human-Robot Interaction (HRI), including Human-Robot Collaboration (HRC): where human and robotic agents work together toward a shared goal. In particular, this behavior is desired in manufacturing tasks for collaborative assembly.

To perform collaborative assembly tasks safely, robots must understand where human operators’ hands are within a shared task space. This requires active safety systems which rely on hand segmentation algorithms to avoid or interact with human collaborators. Despite the plethora of openly available hand segmentation datasets, most do not prioritize operating in hazardous or industrial settings. Rather, these datasets leverage web-sourced data that are biased toward easily accessible objects and environments, e.g., common household items in residential settings \citep{open_x_embodiment_rt_x_2023}. The public datasets that do target industrial domains often suffer from being either: (1) small-scale and lacking human subject diversity \citep{sajedi_uncertainty-assisted_2022, Shilkrot2019WorkingHandsAH}  or (2) generated as synthetic data \citep{grushko_hadr_2023}.

These shortcomings are highly consequential for active safety systems. For instance, most real-time segmentation algorithms leverage convolutional neural networks (CNNs) for hand classification. However, these architectures may over-rely on pixel color values for classification \citep{singh2020assessing}. Thus, underrepresentation in these datasets may unnecessarily imperil people of color.

In an effort to motivate work toward this overlooked task space, we present HAGS: the Hand and Glove Segmentation dataset (\texttt{IRB ID: STUDY00003948}) which is publicly available within our \href{https://dataverse.tdl.org/dataset.xhtml?persistentId=doi:10.18738/T8/85R7KQ}{dataset repository}. This dataset contributes the following:

\begin{enumerate}
    \item The first publicly released human-robot collaboration glovebox hand and glove segmentation dataset. This dataset contains 191 videos of joint assembly tasks totaling 9 hours of content with 1728 frames of pixel-level labels.
    \item The inclusion of ten diverse participants to maximize the relevance of ungloved data. 
    \item Multiclass segmentation for distinguishing between left and right hands as needed by Human-Robot Collaboration applications.
    \item A report on comprehensive baselines for segmentation performance that include metrics for uncertainty quantification, which are largely missing in previous works.
\end{enumerate}

\section{Related Work}
\label{sec:related_work}

There are numerous hand segmentation datasets; however, most focus on aspects of daily life with activities such as cooking or playing cards \citep{bambach_lending_2015, khan_analysis_2018, li_eye_2020}. For a more comprehensive account of these prior works, refer to our Overview of Hand Segmentation Datasets in the Supplemental Material. Although expansive, these datasets are not adequate for situations such as collaborative tasks with robots. Thus, we focus our dataset comparisons on the most relevant works in industrial domains (see Table \ref{tab:industrial_datasets}.)

\begin{table}
\caption{This table summarizes hand segmentation datasets focusing on industrial domain tasks such as part assembly or manipulation. The \textit{Mode} column specifies if color, C, and depth, D, are recorded. \textit{Activity} specifies the robotic dataset's task focus. \textit{Subjects} specifies the number of participants. \textit{Label Method} denotes how the segmentation mask (or ground truth labels) are developed.}
\label{tab:industrial_datasets}
\centering
\resizebox{\columnwidth}{!}{
\begin{tabular}{lllllll}
\toprule
\large
\textbf{Dataset} & \textbf{Mode} & \textbf{Activity} & \textbf{Labels} & \textbf{Subjects} & \textbf{Label Method} & \textbf{\# Classes} \\
\midrule
\normalsize
WorkingHands, 2019 \citep{Shilkrot2019WorkingHandsAH} & CD & Assembly & 7.8k & - & Pixel & 14 \\ 

MECCANO, 2020 \citep{ragusa_meccano_2020} & C & Assembly & - & 20 &  Bound. Box & 21 \\ 

HRC, 2022 \citep{sajedi_uncertainty-assisted_2022} & C & Assembly & 1.3k & 2 & Pixel & 5 \\ 

HaDR, 2023 \citep{grushko_hadr_2023} & CD & Manipulation & 117k & - & Pixel & - \\ 

HAGS$^{*}$ (ours), 2024 \citep{HAGS} & C & Assembly & 1.7k & 10 & Pixel & 2 \\ 
    \bottomrule
\end{tabular}}

\vspace*{-\baselineskip}
\end{table}

\textbf{Task Comparisons.}
Most industrial datasets are not tailored for HRC barring recent work on uncertainty estimation for segmentation tasks pertaining to collaborative assembly \citep{sajedi_uncertainty-assisted_2022}. Unfortunately, this dataset is derived from only two participants. In an effort to expand on this work, we collect additional data on ten participants for two new assembly tasks while incorporating glovebox environment data. Supplementing glovebox data is crucial because this metallic environment presents unique challenges to segmentation such as shine and reflection.

The HaDR dataset \citep{grushko_hadr_2023} features robotic arms and closely resembles our goal to develop robust hand segmentation algorithms. However, no specific HRC task is defined in their dataset and it is entirely synthetic. Similarly, many industrial datasets synthetically augment background color and texture to assess robustness in segmentation models \citep{grushko_hadr_2023, Shilkrot2019WorkingHandsAH}. Additionally, while these datasets contain a plethora of tool and item classes, they do not distinguish right and left-hand information in their labeling, which presents a challenge to extending to future HRI research and applications.

\textbf{Participant Diversity.} 
Well-constructed datasets are crucial to machine learning models' reliability and task performance. In the context of active safety systems, participant diversity in machine learning datasets is of utmost importance. Our work supplements previous industrial datasets \citep{baraldi_gesture_2014, cai_ego-vision_2017, fathi_learning_2011, likitlersuang_interaction_2018, sajedi_uncertainty-assisted_2022} in this regard by including ten diverse participants. Please refer to our datasheet provided in the Supplemental Materials for additional details.

\textbf{Color Invariance.} Several datasets leverage depth data to reduce the influence of lighting and skin coloration on their segmentation models \citep{ grushko_hadr_2023, Shilkrot2019WorkingHandsAH}. However, multi-modal models may still over-rely on RGB features \citep{singh2020assessing}. Other work seeks to completely remove this bias by creating color-agnostic datasets but potentially risks generalizable segmentation performance by forgoing rich RGB signals \citep{kang_hand_2018}. 

Alternative methods used to address color-invariant segmentation include augmenting existing work with simulated data \citep{Shilkrot2019WorkingHandsAH} or generating fully synthetic training datasets \citep{grushko_hadr_2023}. Although most synthetic datasets strive to capture features of real-world data, others aim to build texture and lighting invariance into their models via unrealistic data augmentations \citep{grushko_hadr_2023}.

\textbf{Contributions of HAGS.} Our work addresses the limitations of existing industrial domain datasets by focusing on a previously underserved area: glovebox environments. We present a sizable and diverse dataset comprising real images of ten participants, in contrast to the synthetic images commonly used in other studies. Recognizing the potential application of this technology in active safety systems for human-robot interactions, we have meticulously developed challenging examples using out-of-distribution (OOD) scenarios, such as green screens with distracting images and hands instead of gloves. Furthermore, we encourage the evaluation of uncertainty quantification (UQ) metrics, such as expected calibration error, to enhance safety information. Our findings indicate that prior works do not transfer effectively to our challenging dataset, underscoring the need for further efforts targeting industrial domains.

\section{HAGS Dataset}
\label{sec:HAGS_dataset}

This section provides an overview of the dataset collection, preparation, and annotation processes for HAGS. The Datasheet \citep{gebru_datasheet} provided in the Supplemental Materials provides a more thorough accounting of the dataset. 

\subsection{Data Collection}
Videos are collected in a standard Department of Energy (DOE) glovebox. Two camera angles are provided per video: one 1080p GoPro from a bird's-eye view and one 1080p Intel RealSense Development Kit Camera recording the right side of the participant. Ten participants are included in the study with 16 videos each totaling more than 9 hours of content. Normally distributed frames are collected and annotated from each video, amounting to over 1440 frames. The Unreal Robotics UR3e robot arm, with an attached gripper for handling objects, is used to aid the human subject. The robot is pre-programmed to assist the human participant with sequential, assembly tasks. 

\subsection{Surrogate Joint-Assembly Tasks}

 In order to gather representative data for joint-assembly tasks within a glovebox, two surrogate tasks were designed for human participants to perform. The first task was to assemble a Jenga tower, and the second task was to deconstruct a toolbox.

 The selection of Jenga as a surrogate task supports two critical areas of interest. First, Jenga presents a challenging and generalized pick-and-place assembly problem that is sufficiently representative of high-risk applications. For instance, stacking tasks are essential in biological gloveboxes, where stands and fixtures must be assembled for environmentally controlled experiments; and in manufacturing, where molds need to be assembled before metal casting. Second, the blocks in Jenga, with their elongated and thin shapes with wooden tones are similar to skin tones, making them interesting candidates to include in a challenging hand segmentation dataset. By demonstrating human-robot collaboration using such similar objects in our experiments, we increase confidence that our system can safely function in scenarios where humans and robots collaborate to handle more traditional SNM components.

\textbf{Jenga Task.} Jenga blocks, which loosely resemble the shape, size, and color of human fingers, were chosen to challenge hand segmentation models. The experiment involved three roles: a robot operator, a Jenga block placer, and a participant. The robot operator managed the robot's actions, which included picking up a Jenga block and handing it to the participant for placement. This sequence was repeated until the participant successfully stacked six Jenga blocks.

\textbf{Toolbox Task.} In this experiment, three roles were defined: a robot operator, a tool adjuster, and a participant. The participant was given a closed toolbox secured by four screws. A tool adjuster was on standby while the robot operator oversaw the robot's movements. The robot systematically picked up tools from a stand and handed each one to the participant, who used them to unscrew the toolbox. After using the first two tools, the participant ``rejected'' the selection and handed the third tool back to the robot, which then returned it to the tool adjuster and retrieved the fourth tool for the participant to continue opening the box. Finally, the robot retrieved a previously used and replaced tool from the tool adjuster, providing it to the participant to remove the last screw. Each screwdriver differed in shape and size. The toolbox was white, posing a challenge for segmentation models to distinguish it from the white gloves worn by the participant in the glovebox.

\subsection{Additional Factors.}

Beyond task and participant diversity, two additional factors were altered during the collection process: green screen and glove use.

\begin{figure}[h]
    \centering
    \includegraphics[width=\linewidth]{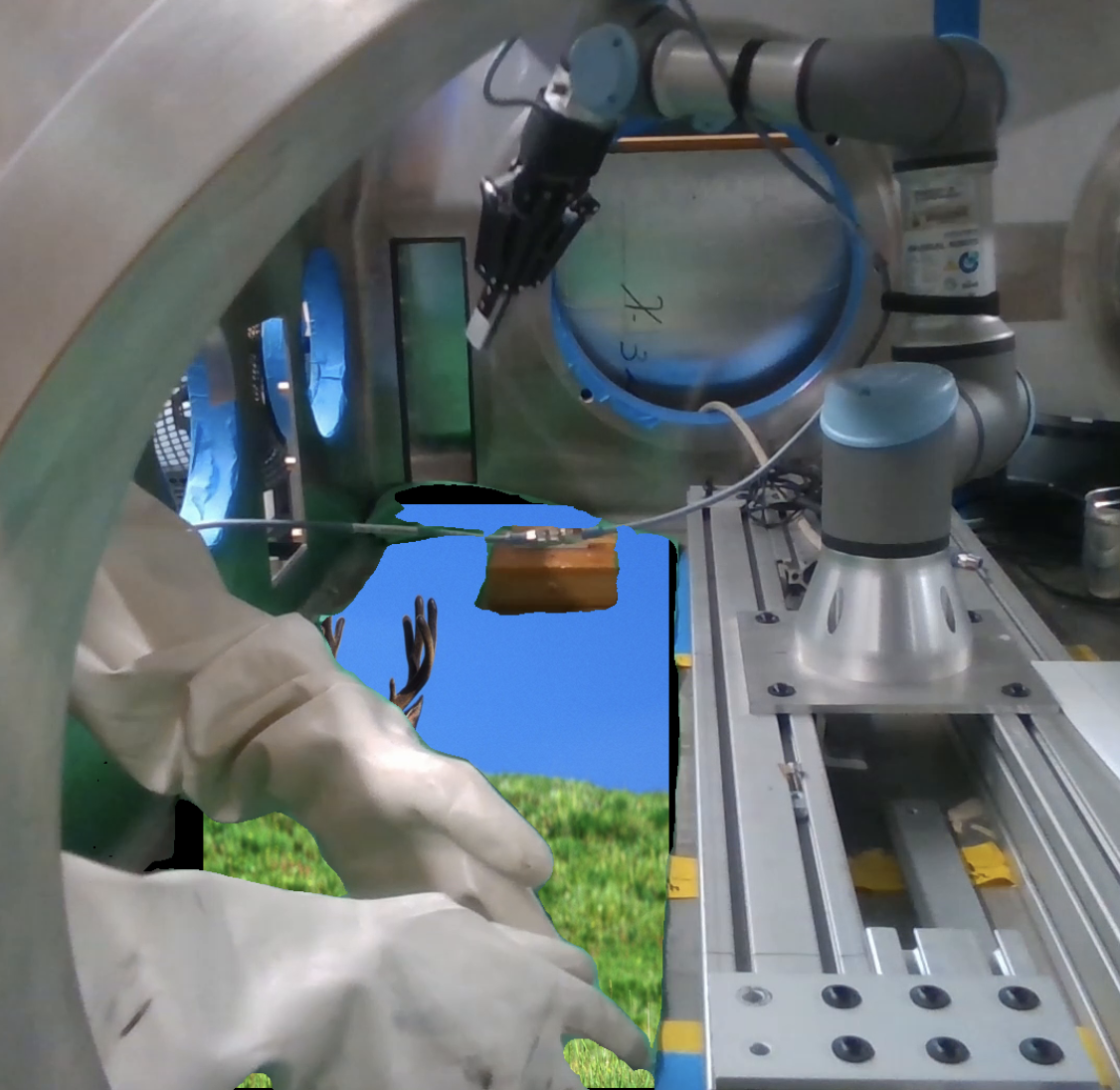}
    \caption{Example of green screen with artificially imposed texture.}
    \label{fig:greenscreen-ex}
\end{figure}

\textbf{Green screen.} A green screen was placed on the bottom of the glovebox for half of the participant videos. This green screen was used to later apply synthetic textures and colors to the background of the real-world images (see Fig. \ref{fig:greenscreen-ex}). The script incorporated in our code repository for generating synthetic images leverages DuckDuckGo Image searches. Notably, these images are not stored within our dataset; rather, we provide the capability for other researchers to independently construct analogous datasets. Frames extracted from videos containing a green screen background are placed into an OOD test set. 

\textbf{Gloves.} Despite operating in a glovebox, half of the recorded videos feature ungloved hands (see the right side of Fig. \ref{fig:gb-intro}.) We incorporated this additional OOD scenario due to the risk of glove tears. Thus, our dataset accounts for rare scenarios where the active safety system must still ensure safe HRC despite skin exposure. These videos are also placed in an OOD test set.

\subsection{Data Preparation}
The combination of the four factors below results in 16 videos per participant. 

\begin{itemize}
    \item Top View / Side View
    \item Toolbox task / Jenga task
    \item Gloves worn / No gloves worn
    \item Green screen included / No green screen included
\end{itemize}

We split sampled frames into an in-distribution (ID) and an OOD set. The ID set contains the most likely glove box operating scenarios. In total, 1440 frames were sampled for labeling. These were equally distributed across all videos, with 120 ID frames and 24 OOD frames sampled per participant.

\subsection{Data Annotation}

Three classes were assigned to each image: left-hand, right-hand, and background. Human annotators were instructed to annotate each hand from the tip to the wrist, and provide their best estimate of the wrist location when the subject was wearing gloves. Using LabelStudio \citep{Label_Studio}, annotators provided key point prompts to MobileSAM \citep{mobile_sam}, a segmentation model which provided a coarse label for annotators to refine. Four of the researchers performed annotations. Two annotators labeled each image to develop inter-annotator agreement (IAA) for label quality. We calculate IAA in two ways: (1) as the average Cohen's Kappa (0.916) and (2) as the average IOU (0.957) between annotator-provided labels across the full dataset , indicating strong agreement. Each frame's annotations were converted to a single PNG file, where the three classes were recorded: left-hand, right-hand, background.

\section{Experiments}
\label{sec:Experiments}

We conducted three experiments to assess the challenging nature of our dataset. \textbf{Experiment A} is a transfer learning experiment designed to demonstrate the limitations of prior works, listed in Table \ref{tab:industrial_datasets}, when applied to our task space. This experiment highlights the lack of sufficient transferability of pretraining on existing datasets to our specific domain. \textbf{Experiment B} studies uncertainty quantification (UQ) involving ID and OOD testing using the HAGS dataset. The OOD frames for all participants include scenarios where gloves are not worn or a green screen is placed in the background. For the green screen frames, we overlay images of different objects to create diverse and challenging testing conditions. This experiment assesses the robustness and reliability of our models in handling diverse, challenging scenarios. In \textbf{Experiment C}, we evaluate the skin tone diversity within industrial hand segmentation datasets using the Monk Skin Tone (MST) scale \citep{Monk_2019}. The objective is to assess the effectiveness of these datasets in addressing the underrepresentation of diverse skin tones in machine learning research, thereby providing insights into their suitability for promoting inclusivity and fairness in model development.

\subsection{Experimental Setup}
For training all models in both experiments, we input 256x256 image sizes and use the Adam optimizer with a learning rate of 1e-3 for the UNet (30 million parameters) and BiSeNetv2 (5 million parameters) architectures, and 8e-4 for MobileSAM (10 million parameters) training. To train or perform inference with MobileSAM, a visual prompt is first required. We adopt a simple approach of selecting a bounding box prompt that contained the whole image. Dropout of p=0.1 was utilized for training of all architectures. We use a variety of data augmentations applied to our training set, including: resize, color jitter, advanced blur, Gaussian noise, and a random rotatation of 90 degrees (p=0.5). 

\textbf{Experiment A.} To train, we use an internal cluster equipped with eight NVIDIA RTX A5000 GPUs. The models: UNet and BiSeNetv2 are trained as three member ensembles and undergo initial pretraining using the WorkingHands (WH) \citep{Shilkrot2019WorkingHandsAH}, HaDr \citep{grushko_hadr_2023}, and HRC \citep{sajedi_uncertainty-assisted_2022} datasets. Following this pretraining phase, we fine-tune the models on varying subsets of our ID HAGS dataset. This approach is designed to test the assertion of prior work that strong performance can be achieved with minimal real-world examples when leveraging their synthetic datasets. We perform the pretraining and fine tuning phases with early stopping on the validation step. We did not include the MECCANO \citep{ragusa_meccano_2020} dataset in our study because it only provides bounding box labels, and our task requires pixel-level segmentation. We evaluate the models on both ID and OOD data from the HAGS dataset. Lastly, the transformer-based model MobileSAM was excluded from this experiment due to its high computational cost for training. This exclusion is not considered a significant limitation, as results from Experiment B demonstrated that its performance, measured by IoU, was substantially lower compared to UNet and BiSeNet when fine-tuned from scratch.

\textbf{Experiment B.} For training, we utilize an internal cluster of three RTX A6000 GPUs. The training is conducted on ID video frames where participants are wearing gloves, and the glovebox background does not include a green screen. We exclude participant 2 from the training set, using their ID frames as the ID testing set. We then evaluate on ID and OOD testing splits. For UNet and MobileSAM we use a batch size of 64, while BiSeNetv2 utilizes a batch size of 128 for training.

\textbf{Experiment C.} This study examines skin tone diversity across existing datasets. The analysis involves the following preprocessing steps: (1) applying white balance, (2) normalizing illumination via histogram equalization, and (3) excluding shadow regions using a shadow threshold hyperparameter to the real world images in each dataset. After preprocessing, we isolate the region of interest corresponding to the ground truth mask for hands and skin. From this region, we compute the average RGB value of the pixels, convert it to a hexadecimal representation, and determine the closest match on the Monk Skin Tone Scale \citep{Monk_2019} (c.f. Figure \ref{fig:mst}) using a minimum color distance metric. The matched tone is then assigned as the classification for that region. However, this method cannot be applied to the HaDr dataset because it does not provide ground truth masks for its real-world images, precluding its inclusion in the analysis. Lastly, for the HAGS dataset, we focus exclusively on the no glove, out-of-distribution (OOD) images because it is most pertinent to the objectives of this study.

\subsection{Results}
\label{sec:Results}
We monitor Intersection over Union (IoU) on test sets as a measure for accuracy, the Expected Calibration Error (ECE) metric for calibration error, and average predictive entropy to analyze model uncertainty. Time per image inference is used to analyze real-time model capabilities. Models are also ensembled for testing to see the impact on metrics. 

\begin{figure*}
    \centering
    \includegraphics[width=0.8\linewidth]{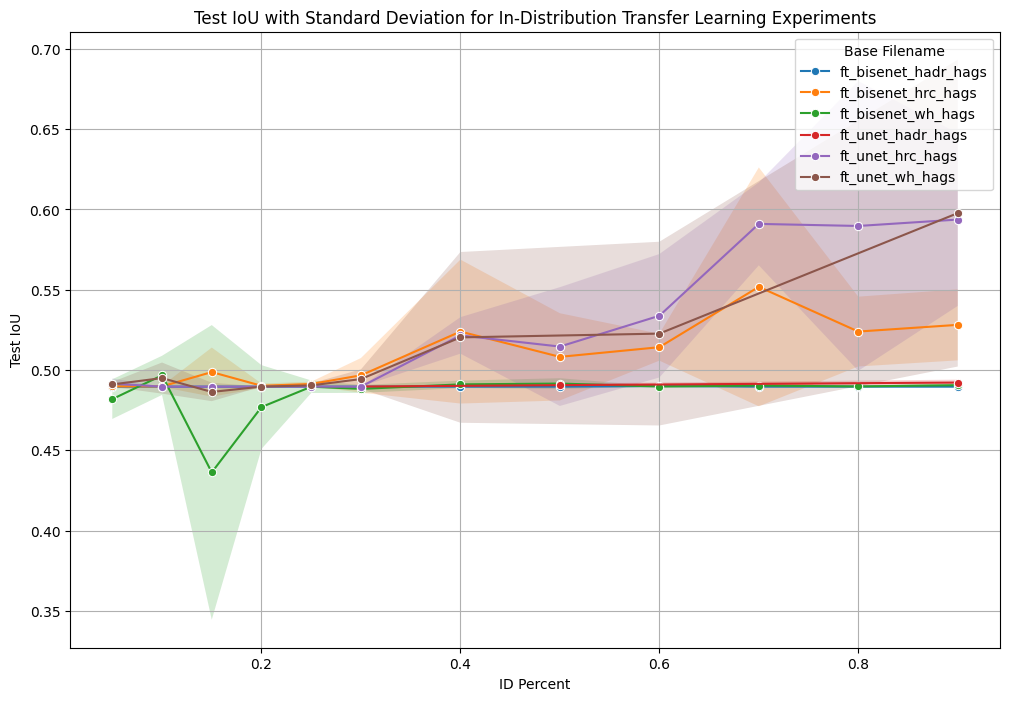}
    \caption{Models were pretrained on HaDr, HRC, or WH datasets then fine-tuned on varying proportions of the HAGS' training dataset. The Intersection-over-Union (IoU) metric was evaluated on the in-distribution portion of the HAGS test set. To read the legend, the \texttt{ft\_} prefix indicates the model was fine-tuned on the in-distribution dataset: HAGS. The rest of the file name follows this convention:\texttt{\{MODEL\_NAME\}\_\{PRETRAINING\_DATASET\}\_\{FINETUNING\_DATASET\}}.}
    \label{fig:tf-learning}
\end{figure*}

\textbf{Experiment A.} As shown in Figure \ref{fig:tf-learning}, the transfer learning experiments presented a significant challenge in all models and datasets evaluated. These challenges are attributed to our focus on real-time applications, which required the use of lower-capacity models. The nature of our task deviates from others, e.g., offline image classification tasks, which may favor the prevailing paradigm of large-scale models and datasets.

Despite notable variance across runs, the HRC dataset emerged as a promising pretraining candidate, particularly in conjunction with the UNet architecture (30 million parameters), which demonstrated superior trainability compared to the BiSeNetv2 model (5 million parameters). However, this variability complicates definitive conclusions regarding the feasibility and necessity of pretraining for our downstream tasks. Specifically, for real-time hand and glove segmentation, the collection of the HRC dataset was essential. No alternative pretraining dataset was able to match the trained-from-scratch fine-tuning performance achieved by the real-time models, underscoring the critical role of task-specific dataset design. Despite the variability, there is some evidence suggesting a modest improvement in downstream task performance when 70\% of the HAGS dataset is incorporated into the fine tuning process. 

The HaDR dataset \citep{grushko_hadr_2023}, the largest pretraining dataset evaluated in this study, produced suboptimal results. This outcome suggests that, given the constraints of small model sizes, curating a smaller, more representative subset of the HaDR dataset could have been more effective than utilizing the dataset in its entirety. However, achieving such a subset requires significant engineering effort, involving extensive trial and error, to identify the most informative samples for the target task.

These findings highlight the nuanced trade-offs between dataset size, model capacity, and downstream performance in the context of real-time applications, where computational constraints play a critical role. Further exploration into dataset engineering and transfer learning stability is warranted to refine these insights.

\begin{table*}[h]
  \caption{Fine-tuning UNet and BiSeNetv2 from scratch on the HAGS ID training set and evaluating OOD performance on the ungloved test split (OOD-Hands). MobileSAM is fine-tuned from its published checkpoint. Reported metric include Intersection over Union (IOU), Expected Calibration Error (ECE), and Predictive Entropy (PE). The reported is the difference between the OOD PE and ID PE: \(\Delta PE = PE_{OOD}-PE_{ID}\).}
  \label{tab:finetuning-hags}
  \centering
  \resizebox{\linewidth}{!}{
  \begin{tabular}{llllllllllll}
    \toprule
    \multicolumn{1}{c}{} & \multicolumn{2}{c}{ID} & \multicolumn{3}{c}{OOD - Hands} & \multicolumn{3}{c}{OOD - Replaced GS} & \multicolumn{3}{c}{OOD - Hands + Replaced GS} \\
    \cmidrule(r){2-3}
    \cmidrule(r){4-6}
    \cmidrule(r){7-9}
    \cmidrule(r){10-12}
    Model     & IOU $\uparrow$ & ECE $\downarrow$ & IOU $\uparrow$ & ECE $\downarrow$ & $\Delta$PE $\uparrow$ & IOU $\uparrow$ & ECE $\downarrow$ & $\Delta$PE $\uparrow$ & IOU $\uparrow$ & ECE $\downarrow$ & $\Delta$PE $\uparrow$ \\
    \midrule
    UNet & 0.8003 & 0.0034 & 0.5663 & 0.0100 & -0.0050 & 0.5628 & 0.0114 & -0.0041 & 0.5694 & 0.0104 & -0.0046 \\
    UNet+dropout & 0.7897 & 0.0036 & 0.5709 & \textbf{0.0095} & -0.0041 & 0.5556 & 0.0115 & -0.0028 & 0.5802 & 0.0098 & -0.0037 \\
    UNet+ensemble & 0.8017 & 0.0043 & \textbf{0.5936} & 0.0100 & -0.0032 & 0.6458 & 0.0095 & -0.0022 & 0.6056 & 0.0099 & -0.0024 \\
    UNet+ensemble+dropout & \textbf{0.8019} & 0.0043 & 0.5753 & 0.0100 & -0.0027 & \textbf{0.6484} & \textbf{0.0092} & -0.0015 & \textbf{0.6125} & \textbf{0.0096} & -0.0028 \\
    \midrule
    BiSeNetv2 & 0.7304 & 0.0052 & 0.5410 & 0.0103 & -0.0040 & 0.5447 & 0.0114 & 0.0016 & 0.5335 & 0.0111 & -0.0021 \\
    BiSeNetv2+dropout & 0.7320 & 0.0059 & 0.5513 & 0.0102 & -0.0001 & 0.5495 & 0.0127 & \textbf{0.0029} & 0.5392 & 0.0116 & \textbf{0.0016} \\
    BiSeNetv2+ensemble & 0.7427 & 0.0056 & 0.5321 & 0.0120 & -0.0038 & 0.5857 & \textbf{0.0109} & -0.0011 & 0.5409 & 0.0119 & -0.0025 \\
    BiSeNetv2+ensemble+dropout & 0.7384 & 0.0058 & 0.4991 & 0.0128 & -0.0037 & 0.5622 & 0.0121 & 0.0017 & 0.5429 & 0.0123 & -0.0027 \\
    \midrule
    MobileSAM & 0.5622 & 0.4020 & 0.5126 & 0.4059 & 0.0000 & 0.5136 & 0.4067 & 0.0002 & 0.5196 & 0.4076 & 0.0004 \\
    MobileSAM+dropout & 0.5486 & 0.4011 & 0.5220 & 0.4057 & 0.0002 & 0.5188 & 0.4065 & 0.0002 & 0.5263 & 0.4074 & 0.0004 \\
    MobileSAM+ensemble & 0.5168 & 0.4012 & 0.5128 & 0.4210 & \textbf{0.0009} & 0.4802 & 0.4171 & 0.0013 & 0.4792 & 0.4070 & 0.0013 \\
    MobileSAM+ensemble+dropout & 0.4974 & 0.4004 & 0.5315 & 0.4223 & 0.0006 & 0.5212 & 0.4193 & 0.0007 & 0.5112 & 0.4086 & 0.0012 \\
    \bottomrule
  \end{tabular}}
\end{table*}

\textbf{Experiment B.} In Table \ref{tab:finetuning-hags}, we present the performance metrics on the ID test set, along with an OOD split focused exclusively on hands. Our analysis reveals that the primary source of variability in IoU scores arises from the choice of evaluation set rather than other factors such as the application of dropout or variations in model architecture. Notably, model ensembling leads to improved IoU scores, particularly evident in the replaced green screen dataset.

\begin{table}
  \caption{Average inference speed reported in seconds and frames per second for the HAGS ID test set.}
  \label{tab:inference-time}
  \centering  
    \resizebox{\columnwidth}{!}{

  \begin{tabular}{lcc}
    \toprule
    Model & Time (s) $\downarrow$ & Frames per Second $\uparrow$ \\
    \midrule
    
    UNet &  \textbf{0.0003} & \textbf{3333} \\
    UNet+ensemble & 0.0073 & 137 \\
    BiSeNetv2 & 0.0008 & 1259 \\
    BiSeNetv2+ensemble & 0.0018 & 56 \\
    MobileSAM & 0.0146 & 68 \\
    MobileSAM+ensemble & 0.0452 & 22 \\
    \bottomrule
  \end{tabular}}
\end{table}

We see low ECE for the ID test set and a progressive increase in ECE for OOD sets, reaching the highest values in the replaced green screen split. When employing model ensembling, we observe the most significant reduction in ECE for the replaced green screen sets. Additionally, our findings indicate that ensembling tends to reduce predictive entropy. Lastly, we observe a counterintuitive pattern where predictive entropy is higher in ID sets compared to OOD sets. We hypothesize that this phenomenon arises because the model frequently misclassifies hands in the OOD set as background with high confidence, resulting in lower entropy for OOD frames.


Table \ref{tab:inference-time} presents the average inference times per image for individual models. As anticipated, CNN-based models achieve real-time inference performance. The transformer model, MobileSAM, while inherently slower, still manages to attain near real-time performance, even when ensembled, at a rate of 22 frames per second.





\begin{figure}[htbp]
    \centering
        \subfigure[Monk Skin Tone (MST) Scale \citep{Monk_2019}.]{
            \includegraphics[width=0.45\textwidth]{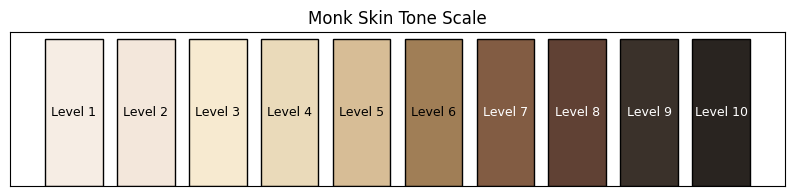}
            \label{fig:mst}
        }
    \hfill
    \subfigure[Histogram in raw counts of MST representation in the real world images from the HAGS \citep{HAGS}, HRC \citep{sajedi_uncertainty-assisted_2022}, and WH \citep{Shilkrot2019WorkingHandsAH} datasets.]{
        \includegraphics[width=0.45\textwidth]{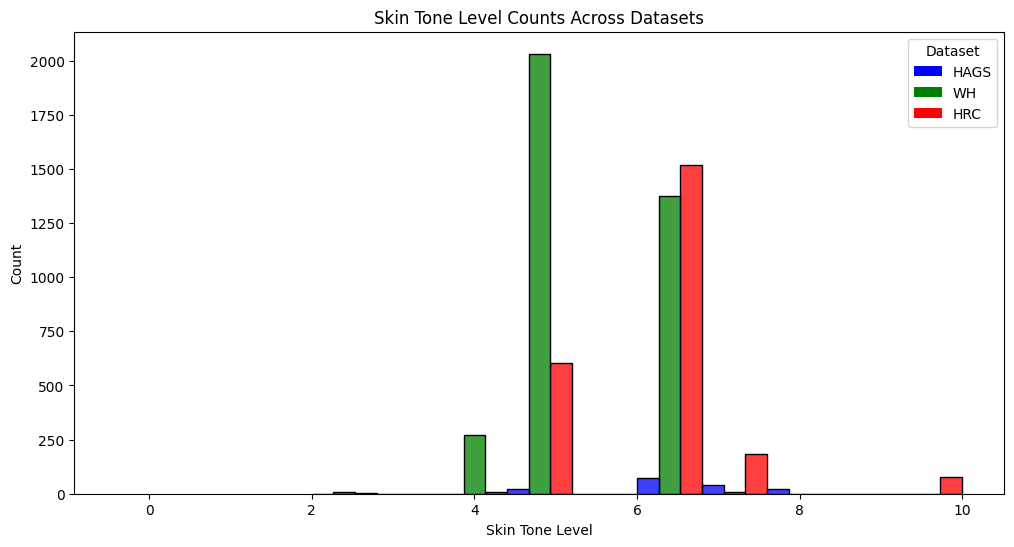}
        \label{fig:raw_mst}
    }
    \hfill
    \subfigure[Normalized histogram in raw counts of MST representation in the real world images from the HAGS \citep{HAGS}, HRC \citep{sajedi_uncertainty-assisted_2022}, and WH \citep{Shilkrot2019WorkingHandsAH} datasets.]{
        \includegraphics[width=0.45\textwidth]{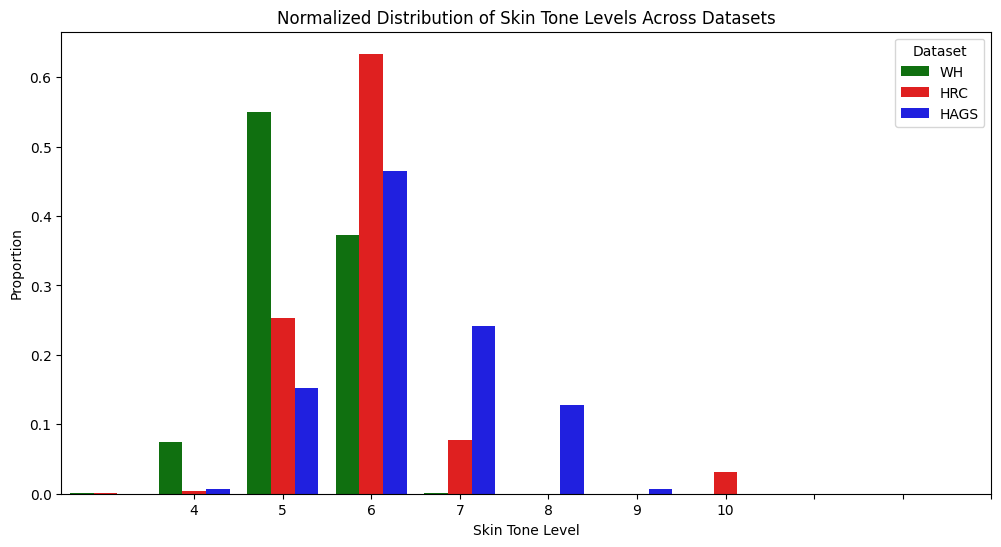}
        \label{fig:normalized_mst}
    }
    \caption{This figure provides a summary of the findings from Experiment C on skin tone representation. Analyzing raw frequency counts, the HRC dataset \citep{sajedi_uncertainty-assisted_2022} demonstrates higher absolute representation of darker skin tones. However, when normalized to proportional representation, the HAGS dataset \citep{HAGS} complements WH \citep{Shilkrot2019WorkingHandsAH} and HRC by exhibiting an increased focus on darker skin tone demographics, addressing gaps in representation evident in the other datasets.}
    \label{fig:mst_investigation}
\end{figure}

\textbf{Experiment C.} The findings of the skin tone diversity investigation suggest that the initial data collection efforts are encouraging, but additional data is required to make a more substantial contribution to this area of study. This is illustrated in Figure \ref{fig:mst_investigation}. Specifically, Figure \ref{fig:normalized_mst} demonstrates that, in proportional terms, the HAGS dataset provides greater coverage of darker skin tones compared to WH and HRC. However, in terms of absolute counts, HRC and WH exhibit broader skin tone representation. These results indicate that rebalancing the skin tone distributions in WH and HRC could address some diversity limitations within these datasets. Nevertheless, the preferred approach remains the collection of more diverse datasets that are representative of real-world use cases, ensuring a broader and more inclusive distribution of skin tones.

\section{Discussion, Conclusions, and Future Work}
\label{sec:discussion_fw}

We present a supervised dataset of real-world RGB image data for hand and glove segmentation, which includes challenging scenarios involving bare hands and green-screened backgrounds. Our experiments demonstrate that pretraining on existing datasets is insufficient to achieve the Intersection over Union (IoU) or uncertainty quantification (UQ) necessary for safe operations in active safety systems for joint assembly, human-robot collaboration tasks. Additionally, training models from scratch on our collected data still presents a challenge when generalized to OOD situations that active safety systems should be better equipped to handle. This underscores the need for further research and specialized datasets in this domain.

\begin{figure}[h]
    \centering
    \includegraphics[width=\linewidth]{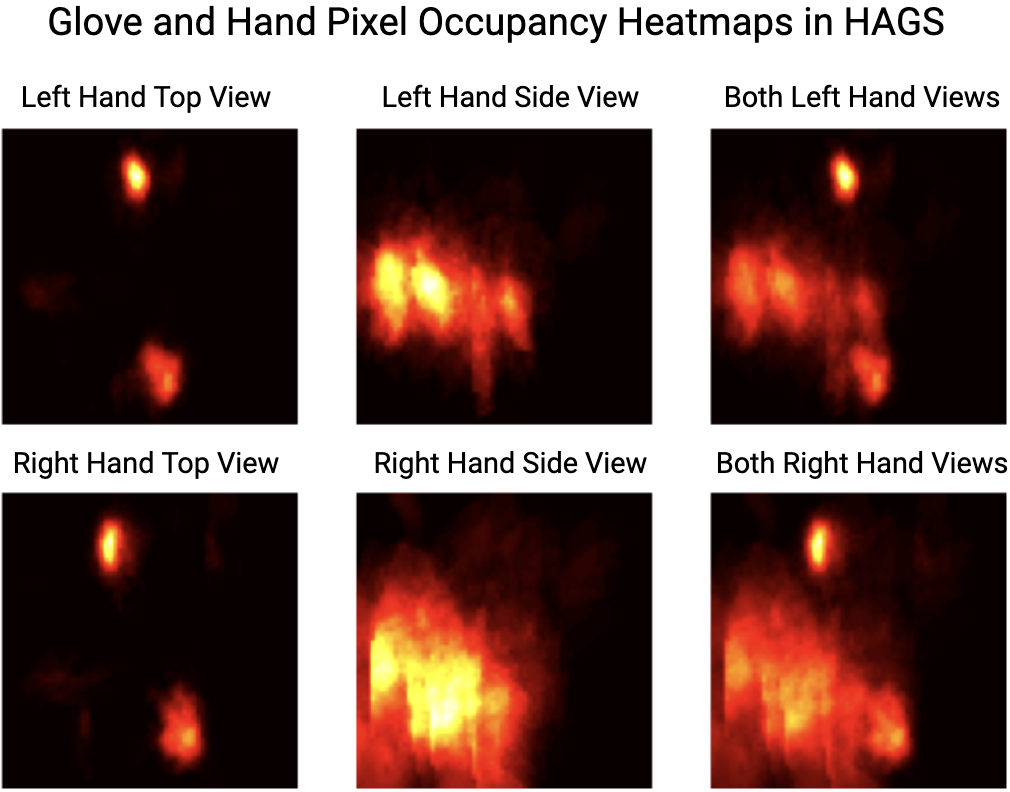}
    \caption{Heat map of hand and glove placements during glovebox tasks in sampled frames.}
    \label{fig:heatmaps}
\end{figure}

There are several areas in which this work can be improved. One significant enhancement would be to increase the diversity and quantity of images and labels, particularly by increasing the training set size to be comparable to other industrial domain datasets. Although the test subjects were diverse, the size of the studies limited the range of diversity that could be included. Future work should further improve diversity with a larger participant pool that includes various age ranges, genders, and skin tones. We currently employ two static camera angles, which, while providing some diversity, resulted in hands being predominantly located in predictable, centered portions of the image, as indicated by the pixel occupancy heat maps (see Figure \ref{fig:heatmaps}). This is a noted, but not significant, limitation since the rigid form of the glovebox limits operators to working in a known and relatively small reachable workspace. The study does not utilize depth sensors to acquire accompanying depth data, nor does it record robot trajectory information. Although both of which could be valuable for advancing other human-robot interaction (HRI) applications, their exclusion simplifies the experimental set-up while mirroring glovebox configurations in the real world. Another limitation is that the participants utilized a single brand of glove. Although gloves are typically similar in color they can vary slightly; and more importantly, they tend to yellow with age. Finally, the method employed for classifying skin tone diversity in Experiment 3 has inherent limitations. While this method may give a general sense of dataset diversity, reliance on basic image processing techniques restricts both the scalability and robustness of the approach. Unfortunately, the Monk Skin Tone (MST) scale \citep{Monk_2019}, while valuable as a standard, lacks an open-source classifier, limiting our analysis.

This work advocates for the application of modern machine learning methods in assisting industrial tasks for underserved use cases. This need is particularly acute for workers whose situation may not be served by typical datasets collected in more common domains. It is the more diverse workforce often employed in these hazardous environments (such as industrial glovebox environments) that can most benefit from the safe use of advanced automation. 

\label{sec:limitations}

\begin{acks}
This manuscript has been approved for unlimited release and has been assigned LA-UR-24-25500 (Rev. 1). This research used resources provided by the Darwin testbed at Los Alamos National Laboratory (LANL) which is funded by the Computational Systems and Software Environments subprogram of LANL’s Advanced Simulation and Computing program (NNSA/DOE). This work was supported by the Laboratory Directed Research and Development program of LANL under project number 20210043DR, C1582/CW8217, and 20250048DR. LANL is operated by Triad National Security, LLC, for the National Nuclear Security Administration of the U.S. Department of Energy (Contract No. 89233218CNA000001). 
\end{acks}

\begin{sm}
\subsection{HAGS Datasheet}
\label{sec:appendix_datasheet}

This section introduced our Hand and Glove Segmentation Dataset (HAGS) datasheet \citep{gebru_datasheet}. The HAGS dataset was collected by the Nuclear and Applied Robotics Group and funded by Los Alamos National Laboratory.

The datasheet is available at \href{https://dataverse.tdl.org/dataset.xhtml?persistentId=doi:10.18738/T8/85R7KQ}{\url{https://dataverse.tdl.org/dataset.xhtml?persistentId=doi:10.18738/T8/85R7KQ}} and its associated software is available at
\href{https://github.com/sanatnair/Glovebox_Segmentation_Dataset_Tools}{\url{https://github.com/sanatnair/Glovebox_Segmentation_Dataset_Tools}} and \href{https://github.com/UTNuclearRoboticsPublic/assembly_glovebox_dataset/tree/main}{\url{https://github.com/UTNuclearRoboticsPublic/assembly_glovebox_dataset/tree/main}}.

\textbf{Hosting, Long Term Preservation, and Maintenance:} The dataset will be maintained by Selma Wanna
(Email: slwanna@utexas.edu) and hosted on the Texas Data Repository: \href{https://dataverse.tdl.org/}{\url{https://dataverse.tdl.org/}}.

\textbf{Statement of Responsibility:}
We accept the responsibility in case of violation of rights, etc., and confirmation of the data license.

\newpage

\large
\noindent\fbox{\parbox{0.95\linewidth}{\centering \textbf{Purpose}}}
\\
\normalsize

This dataset aims to mitigate limitations and biases in existing industrial domain datasets, including low diversity, static environments, and inadequate documentation of HRC datasets and their data collection methods. Additionally, this dataset tackles semantic segmentation of gloves and hands in DOE glovebox environments. We consider the semantic segmentation algorithm to be a component of an active safety system for human-robot collaborative assembly. As such, we construct out-of-distribution (OOD) scenarios to measure task performance and uncertainty quantification to measure algorithmic robustness to potential failures and exaggerated use cases in real-world scenarios. \\

This dataset serves as a foundational resource for training models to enhance working conditions in hazardous environments. It consists of videos capturing two distinct joint assembly tasks with diverse objects, including sampled frames and annotated frames for training and evaluating hand detection models. \\

The dataset consists of 191 videos, totaling approximately 9 hours of footage, along with 1728 annotated in-distribution and out-of-distribution frames, all originating from 10 diverse participants. Furthermore, the dataset emphasizes participants and procedures with various characteristics, such as differing skin tones, arm orientations, camera perspectives, and task execution methods. The two joint assembly tasks within this dataset include the assembly of a block tower and the disassembly of a tool box. In all, the diversity presented enables the dataset to cover a broad range of real-world scenarios. \\

The dataset's final implementation includes the development of real-time baseline models to evaluate the challenging nature of the dataset. Ultimately, with this dataset, we aim to contribute to the ongoing efforts to enhance the safety and accuracy of deep learning models in human-robot collaboration environments. \\



\large
\noindent\fbox{\parbox{0.95\linewidth}{\centering \textbf{Motivation}}}
\normalsize
\\
\\
This dataset was created to enable research on real-time hand segmentation for Human Robot Collaboration (HRC) tasks in industrial domains. The focus is on joint assembly of objects in glovebox environments. We aim to mitigate limitations and biases in existing industrial domain datasets, including low diversity, static environments, and inadequate documentation of HRC datasets and their data collection methods. Because we view the real-time segmentation algorithm as an active safety-system, we construct out-of-distribution (OOD) scenarios to measure task performance and uncertainty quantification. \\

The dataset was created by Shivansh Sharma, Mathew Huang, Sanat Nair, Alan Wen, Christina Petloway, Juston Moore, Selma Wanna, and Mitch Pryor at the University of Texas at Austin. \\

The funding of the dataset is from Los Alamos National Laboratory in the form of a Laboratory Directed Research and Development program (20210043DR) and contract C1582/CW8217.\\

\large
\noindent\fbox{\parbox{0.95\linewidth}{\centering \textbf{Composition of the Dataset}}}
\normalsize
\\
\\
The instances that comprise the dataset  are videos of test subject performing assembly tasks and sampled images from those videos. Additionally, there are supervised labels provided for each sampled image by two annotators. \\

There are 16 videos per test subject, from which sampled frames and annotations were created over 2 experiments, 4 variables, 2 camera angles, and 10 subjects. \\

The dataset includes 10 diverse participants and is self-contained; it does not include external sources or assets. \\

There are 1,728 in distribution (ID) and OOD frames total. \\

The dataset is a sample of instances. No tests were run to determine how representative the frames are; however, pixel heatmaps provided in Figure \ref{fig:heatmaps} indicate bias in hand location. Finer sampling should result in a more diverse dataset. \\


There are labels provided only for the images we have sampled. It is possible for other users to sample more images using the provided script: \href{https://github.com/sanatnair/Glovebox_Segmentation_Dataset_Tools}{\url{https://github.com/sanatnair/Glovebox_Segmentation_Dataset_Tools}}. Thereafter, software like LabelStudio \citep{Label_Studio} can assist in creating the new images' ground truths. \\

Each label is a pixel-wise map of three classes: background, left hand, or right hand. These labels and images are saved in PNG formats. \\

There is information missing from two test subject participants. We retained their corrupted files but removed their data from the final dataset. \\

There are no recommended data splits for the ID dataset, i.e., images which contain gloved hands and no green screen. However, for the OOD portion, i.e., images which contain either ungloved hands or green screens with and without overlaid images, we strongly recommend to use this data only for testing sets. \\

Three instances of noise are identified in this dataset after quality inspection. Firstly, it is observed that only 22 OOD frames were sampled with Participant 6, as opposed to the expected 24. This discrepancy occurs specifically within the Side\textunderscore View of the Jenga\textunderscore task. Secondly, there is an issue with the orientation of the Top\textunderscore View $>$ Toolbox\textunderscore GL video and its subsequent frames and annotations for Participant 3. Finally, the cropped video included in the dataset for Participant 3 $>$ Side\textunderscore View $>$ Jenga\textunderscore GL\textunderscore G is not the same as the video used for generating sampled frames and annotations.  \\

The dataset is self-contained barring the instances with green screens with overlaid images. These images were sourced from Duck Duck Go images using a script: \href{https://github.com/UTNuclearRoboticsPublic/assembly_glovebox_dataset/blob/main/data/convert_scripts/replace_green.py}{\url{https://github.com/UTNuclearRoboticsPublic/assembly_glovebox_dataset/blob/main/data/convert_scripts/replace_green.py}}. There is no guarantee that the specific images will remain constant over time. There is not an archival version of this portion of the dataset that we can distribute. \\

No confidential, sensitive, anxiety inducing, or offensive images are contained in the dataset. Our dataset does not identify subpopulations. \\

It is possible but unlikely to identify individuals from the datasets. \\

\large
\noindent\fbox{\parbox{0.95\linewidth}{\centering \textbf{Data Collection}}}
\normalsize
\\

The following list contains the hardware and software used in the data acquisition:\\

\begin{itemize}
    \item GoPro Hero 7 and a RealSense Development Kit Camera SR300 
    \item 1080p, 30 fps, 0.5 Ultrawide Lens.
    \item UR3e from Universal Robots, programmed using the pendant tablet.
    \item Experiments conducted in the same location, with the same robot and glovebox.
\end{itemize}

A deterministic sampling strategy was used. \\

Students and researchers from the University of Texas at Austin were involved in this study. They were not paid, but were entered into a raffle for an Amazon Gift Card.\\

An ethical review process was conducted. An IRB exemption was granted for this study. This information is viewable online at \href{https://dataverse.tdl.org/file.xhtml?fileId=599918&version=1.0}{\url{https://dataverse.tdl.org/file.xhtml?fileId=599918&version=1.0}}. \\

We collected the data directly from the individual participants. The data was collected from Fall 2022 - Spring 2023.\\

All participants received an informed consent form: \href{https://dataverse.tdl.org/file.xhtml?fileId=599917&version=1.0}{\url{https://dataverse.tdl.org/file.xhtml?fileId=599917&version=1.0}} and were aware of our study and data collection.\\

All participants consented to the collection of their data. However, participants can revoke their data by emailing the study organizer.\\

\large
\noindent\fbox{\parbox{0.95\linewidth}{\centering \textbf{Data Processing}}}
\normalsize
\\

Removal of instances and processing of missing values was done for two participants. This was due to a data corruption of our hard drive making their runs unrecoverable. \\

The raw data is saved in addition to the preprocessed data because we retain the video files in our dataset. \\

The software for labeling the data is available from LabelStudio \citep{Label_Studio}. \\

Frames are labeled as left and right hand with LabelStudio software \citep{Label_Studio} and the MobileSAM \citep{mobile_sam} backend. Models were trained on frames compressed to 256x256 px. Ground truth masks and original footage data is available, while raw unprocessed data is private. \\

\large
\noindent\fbox{\parbox{0.95\linewidth}{\centering \textbf{Ethics}}}
\normalsize
\\
\\
Participants with recognizable features are asked to cover them up through the use of makeup or other methods to prevent identifiability. \\

The only characteristic identified by the dataset is race. \\

The experimental procedure used to collect the data is reviewed by the internal review board of the University of Texas at Austin (\texttt{IRB ID: STUDY00003948}). \\

Individuals are allowed to revoke their consent. \\

\large
\noindent\fbox{\parbox{0.95\linewidth}{\centering \textbf{Uses}}}
\normalsize
\\
\\
At the time of publication, this dataset has only been used in our original work.\\

This repository: \href{https://github.com/UTNuclearRoboticsPublic/assembly_glovebox_dataset/tree/main}{\url{https://github.com/UTNuclearRoboticsPublic/assembly_glovebox_dataset/tree/main}} links to our public repository, which will contain a running list of associated papers or publications.\\

This dataset could potentially be used for intention recognition in Human Robot Interaction research. An example use case could be classifying future tool use/need based on user behavior. However, our dataset could be improved for this use case with gesture annotations.\\

Future uses could be impacted because we did not record robot trajectories when developing this dataset. Additionally, we did not use depth sensors or RGB-D cameras for data collection. \\

This dataset was collected in a DOE glovebox and as such the data should not be used on its own to develop segmentation algorithms for markedly different scenarios. We note that generalization in niche industrial domains is a current problem in ML, so please proceed with caution if you plan to adapt this work to your use case. \\

The data may be impacted by future changes in demographics and the proportion of representation of groups may change. The dataset is unfit for applications outside of the glovebox and applications not related to hands. \\

\large
\noindent\fbox{\parbox{0.95\linewidth}{\centering \textbf{Distribution}}}
\normalsize
\\
\\
We plan to distribute this dataset to DOE laboratories. \\

The dataset is hosted via the Texas Data Repository: \href{https://dataverse.tdl.org/dataset.xhtml?persistentId=doi:10.18738/T8/85R7KQ}{\url{https://dataverse.tdl.org/dataset.xhtml?persistentId=doi:10.18738/T8/85R7KQ}}. Supporting GitHub repositories include \href{https://github.com/UTNuclearRoboticsPublic/assembly_glovebox_dataset/tree/main}{\url{https://github.com/UTNuclearRoboticsPublic/assembly_glovebox_dataset/tree/main}} and \href{https://github.com/sanatnair/Glovebox_Segmentation_Dataset_Tools}{\url{https://github.com/sanatnair/Glovebox_Segmentation_Dataset_Tools}}. \\

The dataset was released February 29, 2024. \\

This dataset is available under a CC0 1.0 license. Both code repositories are released under BSD 3.0 licenses.\\

No third parties imposed IP-based or other restrictions on the data associated with the instances.\\

This dataset does not violate export controls or other regulatory restrictions. \\


\newpage

\subsection{Overview of Hand Segmentation Datasets}
\label{sec:appendix_megatable}

\begin{table}[h]
\caption{Domain and task overview of hand segmentation datasets.}
\label{tab:app1}
\centering
\resizebox{\columnwidth}{!}{
\begin{tabular}{lccccc}
\toprule

\large
\textbf{Dataset} & \textbf{Year} & \textbf{Mode} & \textbf{Device} & \textbf{Type of Activity} & \textbf{Setting} \\ \midrule
\normalsize
\textbf{GTEA \citep{fathi_learning_2011}} & 2011 & C & GoPro & Cooking & Kitchen \\ \midrule
\textbf{ADL \citep{pirsiavash_detecting_2012}} & 2012 & C & GoPro & Daily Life & Home \\ \midrule
\textbf{EDSH \citep{li_pixel-level_2013}} & 2013 & C & - & Daily Life & Home / Outdoors \\ \midrule
\textbf{Interactive Museum \citep{baraldi_gesture_2014}} & 2014 & C & - & Gesture & Museum \\ \midrule
\textbf{EgoHands \citep{bambach_lending_2015}} & 2015 & C & Google Glass & Manipulation & Social Setting \\ \midrule
\textbf{Maramotti \citep{baraldi_gesture_2015}} & 2015 & C & - & Gesture & Museum \\ \midrule
\textbf{UNIGE Hands \citep{betancourt_dynamic_2015}} & 2015 & C & GoPro Hero3+ & Daily Life & Various Daily Locations \\ \midrule
\textbf{GUN-71 \citep{rogez_understanding_2015}} & 2015 & CD & Creative Senz3D & Daily Life & Home \\ \midrule
\textbf{RGBD Egocentric Action \citep{wan_mining_2015}} & 2015 & CD & Creative Senz3D & Daily Life & Home \\ \midrule
\textbf{Fingerwriting in mid-air \citep{chang_spatio-temporal_2016}} & 2016 & CD & Creative Senz3D & Writing & Office \\ \midrule
\textbf{Ego-Finger \citep{huang_pointing_2016}} & 2016 & C & - & Gesture & Various Daily Locations \\ \midrule
\textbf{ANS able-bodied \citep{likitlersuang_interaction_2018}} & 2016 & C & Looxie 2 & Daily Life & Home \\ \midrule
\textbf{UT Grasp \citep{cai_ego-vision_2017}} & 2016 & C & GoPro Hero2 & Manipulation & - \\ \midrule
\textbf{GestureAR \citep{mohatta_robust_2017}} & 2017 & C & Nexus 6 and Moto G3 & Gesture & Various Backgrounds \\ \midrule
\textbf{EgoGesture \citep{wu_yolse_2017}} & 2017 & C & - & Gesture & - \\ \midrule
\textbf{Egocentric hand-action \citep{xu_hand_2016}} & 2017 & D & Softkinetic DS325 & Gesture & Office \\ \midrule
\textbf{BigHand2.2M \citep{yuan_bighand22m_2017}} & 2017 & D & Intel RealSense SR300 & Gesture/Pose & - \\ \midrule
\textbf{Desktop Action \citep{cai_desktop_2019}} & 2018 & C & GoPro Hero 2 & Daily Life & Office \\ \midrule
\textbf{Epic Kitchens \citep{damen_scaling_2018}} & 2018 & C & GoPro & Cooking & Kitchen \\ \midrule
\textbf{FPHA \citep{garcia-hernando_first-person_2018}} & 2018 & CD & Intel RealSense SR300 & Gesture / Pose / Daily life & Home \\ \midrule
\textbf{EYTH \citep{khan_analysis_2018}} & 2018 & C & - & - & - \\ \midrule
\textbf{HandOverFace \citep{khan_analysis_2018}} & 2018 & C & - & - & - \\ \midrule
\textbf{EGTEA+  \citep{li_eye_2020}} & 2018 & C & SMI wearable eye-tracker & Cooking & Kitchen \\ \midrule
\textbf{THU-READ  \citep{tang_multi-stream_2019}} & 2018 & CD & Primesense Carmine & Daily Life & Home \\ \midrule
\textbf{EgoGesture \citep{zhang_egogesture_2018}} & 2018 & CD & Intel RealSense SR300 & Gesture & - \\ \midrule
\textbf{HOI \citep{kang_hand_2018}} & 2018 & D & Kinect V2 & Manipulation & - \\ \midrule
\textbf{EgoDaily \citep{cruz_is_2019}} & 2019 & C & GoPro Hero5 & Daily Life & Various Daily Locations \\ \midrule
\textbf{ANS SCI \citep{likitlersuang_egocentric_2019}} & 2019 & C & GoPro Hero4 & Daily Life & Home \\ \midrule
\textbf{KBH \citep{wang_recurrent_2019}} & 2019 & C & HTC Vive & Manipulation / Keyboard & Office \\ \midrule
\textbf{WorkingHands \citep{Shilkrot2019WorkingHandsAH}} & 2019 & CD & Kinect V2 & Assembly & Industrial \\ \midrule
\textbf{Freihand \citep{zimmermann_freihand_2019}} & 2019 & C & - & Grasp / Pose & Green Screen and Outdoors \\ \midrule
\textbf{MECCANO \citep{ragusa_meccano_2020}} & 2020 & C & Intel RealSense SR300 & Assembly & Desk \\ \midrule
\textbf{HRC \citep{sajedi_uncertainty-assisted_2022}} & 2022 & C & iPhone 11 Pro & Assembly & Industrial \\ \midrule
\textbf{HaDR \citep{grushko_hadr_2023}} & 2023 & CD & Realsense L515 & Manipulation & Industrial \\ \midrule
\textbf{Our Dataset} & 2024 & C & Intel Realsense SR300 and GoPro Hero 7 & Assembly & Industrial (glovebox) \\ \bottomrule
\end{tabular}}
\end{table}

\begin{table}[h]

\caption{Overview of the data-specific qualities of hand segmentation datasets. This table focuses on the number of collected, labeled frames and number of subjects which participated in the data collection.}
\label{tab:quantit-overview}
\resizebox{\columnwidth}{!}{

\begin{tabular}{lccccccc}
\toprule
\large
\textbf{Dataset} 
&  \textbf{Frames} 
&\textbf{Annotated Frames} & \textbf{Videos} & \textbf{Duration} & \textbf{Subjects} & \textbf{Resolution} & \textbf{Annotation} \\  \midrule
\normalsize
\textbf{GTEA \citep{fathi_learning_2011}} 
&  ~31K 
&663 & 28 & 34 m & 4 & 1280 x 720 & act msk \\ \midrule
\textbf{ADL \citep{pirsiavash_detecting_2012}} 
&  >1M 
&1,000,000 & 20 & ~10 h & 20 & 1280 x 720 & act obj \\ \midrule
\textbf{EDSH \citep{li_pixel-level_2013}} 
&  ~20K 
&442 & 3 & ~10 m & - & 1280 x 720 & msk \\ \midrule
\textbf{Interactive Museum \citep{baraldi_gesture_2014}} 
&   - 
&& 700 & - & 5 & 800 x 450 & gst msk \\ \midrule
\textbf{EgoHands \citep{bambach_lending_2015}} 
&  ~130K 
&4800 & 48 & 72 m & 8 & 1280 x 720 & msk \\ \midrule
\textbf{Maramotti \citep{baraldi_gesture_2015}} 
&  - 
&- & 700 & - & 5 & 800 x 450 & gst msk \\ \midrule
\textbf{UNIGE Hands \citep{betancourt_dynamic_2015}} 
&  ~150K 
&- & - & 98 m & - & 1280 x 720 & det \\ \midrule
\textbf{GUN-71 \citep{rogez_understanding_2015}} 
&  ~12K 
&9100 & - & - & 8 & - & grs \\ \midrule
\textbf{RGBD Egocentric Action \citep{wan_mining_2015}} 
&  - 
&- & - & - & 20 & C: 640x480 D: 320 x 240 & act \\ \midrule
\textbf{Fingerwriting in mid-air \citep{chang_spatio-temporal_2016}} 
&  ~8K 
&2500 & - & - & - & - & ftp gst \\ \midrule
\textbf{Ego-Finger \citep{huang_pointing_2016}} 
&  ~93K 
&- & 24 & - & - & 640 x 480 & det ftp \\ \midrule
\textbf{ANS able-bodied \citep{likitlersuang_interaction_2018}} 
&  - 
&- & - & 44 m & 4 & 640 x 480 & det ftp \\ \midrule
\textbf{UT Grasp \citep{cai_ego-vision_2017}} 
&  - 
&- & 50 & ~4 h & 5 & 960 x 540 & grs \\ \midrule
\textbf{GestureAR \citep{mohatta_robust_2017}} 
&  ~51K 
&- & 100 & - & 8 & 1280 x 720 & gst \\ \midrule
\textbf{EgoGesture \citep{wu_yolse_2017}} 
&  ~59K 
&- & - & - & - & - & det ftp gst \\ \midrule
\textbf{Egocentric hand-action \citep{xu_hand_2016}} 
&  ~154K 
&- & 300 & - & 26 & 320 x 240 & gst \\ \midrule
\textbf{BigHand2.2M \citep{yuan_bighand22m_2017}} 
&  ~290K 
&~290K & - & - & - & 640 x 480 & pos \\ \midrule
\textbf{Desktop Action \citep{cai_desktop_2019}} 
&  ~324K 
&660 & 60 & 3 h & 6 & 1920 x 1080 & act msk \\ \midrule
\textbf{Epic Kitchens \citep{damen_scaling_2018}} 
&  11.5M 
&- & - & 55h & 32 & 1920 x 1080 & act \\ \midrule
\textbf{FPHA \citep{garcia-hernando_first-person_2018}} 
&  ~100K 
&~100K & 1175 & - & 6 & C: 1920 x 1080 D: 640 x 480 & act pos \\ \midrule
\textbf{EYTH \citep{khan_analysis_2018}} 
&  1290 
&1290 & 3 & - & - & 384 x 216 & msk \\ \midrule
\textbf{HandOverFace \citep{khan_analysis_2018}} 
&  300 
&300 & - & - & - & 384 x 216 & msk \\ \midrule
\textbf{EGTEA+  \citep{li_eye_2020}} 
&  >3M 
&~15k & 86 & ~28h & 32 & 1280 x 960 & act gaz msk \\ \midrule
\textbf{THU-READ  \citep{tang_multi-stream_2019}} 
&  ~343K 
&652 & 1920 & - & 50 & 640 x 480 & act msk \\ \midrule
\textbf{EgoGesture \citep{zhang_egogesture_2018}} 
&  ~3M 
&- & 24161 & - & 50 & 640 x 480 & gst \\ \midrule
\textbf{HOI \citep{kang_hand_2018}} 
&  27525 
&- & - & - & 6 & - & msk \\ \midrule
\textbf{EgoDaily \citep{cruz_is_2019}} 
&  ~50K 
&~50K & 50 & - & 10 & 1920 x 1080 & det hid \\ \midrule
\textbf{ANS SCI \citep{likitlersuang_egocentric_2019}} 
&  - 
&~33K & - & - & 17 & 480 x 854 & det int \\ \midrule
\textbf{KBH \citep{wang_recurrent_2019}} 
&  ~12.5K 
&~12.5K & 161 & - & 50 & 230 x 306 & msk \\ \midrule
\textbf{WorkingHands \citep{Shilkrot2019WorkingHandsAH}} 
&  4.2K syn, 3.7K Real 
&~7.8k & 39 & - & - & 1920 × 1080 & obj msk \\ \midrule
\textbf{Freihand \citep{zimmermann_freihand_2019}} 
&  37000 
&37000 & - & - & 32 & - & pos \\ \midrule
\textbf{MECCANO \citep{ragusa_meccano_2020}} 
&  ~299K 
&- & 20 & 21 min & 20 & 1280 x 720 & act obj msk \\ \midrule
\textbf{HRC \citep{sajedi_uncertainty-assisted_2022}} 
&  598 
&- & 13 & - & 2 & 380 x 180 (resized) & msk \\ \midrule
\textbf{HaDR \citep{grushko_hadr_2023}} 
&  117000 
&117000 & - & - & - & 640 x 480 & msk \\ \midrule
\textbf{Our Dataset} &  4320 &2880 & 160 & 8 h & 10 & 64 x 64 (resized) & msk \\ \bottomrule
\end{tabular}}
\end{table}

\begin{table}[h]
\caption{This table highlights the kinds of labeling provided for hand segmentation datasets. We also consider if that majority of the reported data is synthetic or real.}
\label{tab:synth-v-real}
\resizebox{\columnwidth}{!}{
\begin{tabular}{lcccc}
\toprule
\large
\textbf{Dataset} &  \textbf{Method} 
&\textbf{Simulated data} & \textbf{Total Classes} & \textbf{Non Hand Object Classes} \\  \midrule
\normalsize
\textbf{GTEA \citep{fathi_learning_2011}} &  Pixel 
&N & 17 & 16 \\ \midrule
\textbf{ADL \citep{pirsiavash_detecting_2012}} 
&  Bounding Box 
&N & 11 & 10 \\ \midrule
\textbf{EDSH \citep{li_pixel-level_2013}} 
&  Pixel 
&N & 1 & 0 \\ \midrule
\textbf{Interactive Museum \citep{baraldi_gesture_2014}} &  Bounding Box 
&N & 1 & 0 \\ \midrule
\textbf{EgoHands \citep{bambach_lending_2015}} 
&  Pixel 
&N & 4 & 0 \\ \midrule
\textbf{Maramotti \citep{baraldi_gesture_2015}} 
&  Pixel 
&N & 1 & 0 \\ \midrule
\textbf{UNIGE Hands \citep{betancourt_dynamic_2015}} 
&  - 
&N & 1 & 0 \\ \midrule
\textbf{GUN-71 \citep{rogez_understanding_2015}} 
&  Key Points + Forces 
&N & 1 & 0 \\ \midrule
\textbf{RGBD Egocentric Action \citep{wan_mining_2015}} 
&  - 
&N & 1 & 0 \\ \midrule
\textbf{Fingerwriting in mid-air \citep{chang_spatio-temporal_2016}} 
&  Key Point 
&N & 1 & 0 \\ \midrule
\textbf{Ego-Finger \citep{huang_pointing_2016}} 
&  Bounding Box + Points 
&N & 1 & 0 \\ \midrule
\textbf{ANS able-bodied \citep{likitlersuang_interaction_2018}} 
&  Bounding Box 
&N & 1 & 0 \\ \midrule
\textbf{UT Grasp \citep{cai_ego-vision_2017}} 
&  Bounding Box 
&N & 1 & 0 \\ \midrule
\textbf{GestureAR \citep{mohatta_robust_2017}} 
&   Pixel 
&& 1 & 0 \\ \midrule
\textbf{EgoGesture \citep{wu_yolse_2017}} 
&  Bounding Box + Points 
&N & 1 & 0 \\ \midrule
\textbf{Egocentric hand-action \citep{xu_hand_2016}} 
&  Bounding Box 
&Y & 1 & 0 \\ \midrule
\textbf{BigHand2.2M \citep{yuan_bighand22m_2017}} 
&  Key Point 
&N & 1 & 0 \\ \midrule
\textbf{Desktop Action \citep{cai_desktop_2019}} 
&  Pixel 
&N & 1 & 0 \\ \midrule
\textbf{Epic Kitchens \citep{damen_scaling_2018}} 
&  Bounding Box 
&N & 324 & 323 \\ \midrule
\textbf{FPHA \citep{garcia-hernando_first-person_2018}} 
&  Key Point 
&N & 1 & 0 \\ \midrule
\textbf{EYTH \citep{khan_analysis_2018}} 
&  Pixel 
&N & 1 & 0 \\ \midrule
\textbf{HandOverFace \citep{khan_analysis_2018}} 
&  Pixel 
&N & 1 & 0 \\ \midrule
\textbf{EGTEA+  \citep{li_eye_2020}} 
&  Pixel 
&N & 1 & 0 \\ \midrule
\textbf{THU-READ  \citep{tang_multi-stream_2019}} 
&  Pixel 
&N & 1 & 0 \\ \midrule
\textbf{EgoGesture \citep{zhang_egogesture_2018}} 
&  Pixel 
&N & 1 & 0 \\ \midrule
\textbf{HOI \citep{kang_hand_2018}} 
&  Bounding Box 
&N & 1 & 0 \\ \midrule
\textbf{EgoDaily \citep{cruz_is_2019}} 
&  Bounding Box 
&N & 1 & 0 \\ \midrule
\textbf{ANS SCI \citep{likitlersuang_egocentric_2019}} 
&  Pixel 
&N & 1 & 0 \\ \midrule
\textbf{KBH \citep{wang_recurrent_2019}} 
&  Pixel 
&N & 1 & 0 \\ \midrule
\textbf{WorkingHands \citep{Shilkrot2019WorkingHandsAH}} 
&  Pixel 
&Y & 14 & 13 \\ \midrule
\textbf{Freihand \citep{zimmermann_freihand_2019}} 
&  Key Point 
&N & 1 & 0 \\ \midrule
\textbf{MECCANO \citep{ragusa_meccano_2020}} 
&  Bounding Box 
&N & 21 & 20 \\ \midrule
\textbf{HRC \citep{sajedi_uncertainty-assisted_2022}} 
&  Pixel 
&N & 5 & 1 \\ \midrule
\textbf{HaDR \citep{grushko_hadr_2023}} 
&  Pixel 
&Y & 1 & 0 \\ \midrule
\textbf{Our Dataset} &  Pixel &N & 2 & 0 \\ \bottomrule
\end{tabular}}
\end{table}
\end{sm}

\bibliographystyle{SageH}
\bibliography{references}

\end{document}